\documentclass[conference]{IEEEtran}
\IEEEoverridecommandlockouts
\usepackage{cite}
\usepackage{amsmath,amssymb,amsfonts}
\usepackage{graphicx}
\usepackage{textcomp}
\usepackage{xcolor}
\usepackage{amsmath,amsfonts}
\usepackage{algorithm}
\usepackage{array}
\usepackage[caption=false,font=normalsize,labelfont=sf,textfont=sf]{subfig}
\usepackage{textcomp}
\usepackage{stfloats}
\usepackage{url}
\usepackage{verbatim}
\usepackage{graphicx}
\usepackage{hyperref}
\usepackage{cite}
\usepackage{algpseudocode}
\usepackage{paralist}
\usepackage{float}
\usepackage{booktabs}
\usepackage{amssymb}
\usepackage{pifont}
\newcommand{\cmark}{\ding{51}} 
\newcommand{\xmark}{\ding{55}} 
\usepackage{listings}

\usepackage{amsmath}        
\usepackage{siunitx}        
\usepackage{xcolor}         
\usepackage{hyperref}       
\usepackage{enumitem}       

\def\BibTeX{{\rm B\kern-.05em{\sc i\kern-.025em b}\kern-.08em
    T\kern-.1667em\lower.7ex\hbox{E}\kern-.125emX}}
\begin{document}

\title{ConvoyNext: A Scalable Testbed Platform for Cooperative Autonomous Vehicle Systems\\
\thanks{This work was partially supported by the National Science Foundation
under Grant CNS-1932037. The code and supplementary material are available
at: https://github.com/Hmaghsoumi/ConvoyNext.}
}

\author{\IEEEauthorblockN{Hossein Maghsoumi, Yaser Fallah}
\IEEEauthorblockA{\textit{Department of Electrical and Computer Engineering} \\
\textit{University of Central Florida}\\
Orlando, USA \\
hossein.maghsoumi@ucf.edu, yaser.fallah@ucf.edu}
}

\maketitle

\IEEEpubid{%
  \begin{minipage}{\textwidth}
    \vspace{6.8\baselineskip}   
    \centering
    \fbox{%
      \parbox{0.92\textwidth}{\centering\small
        This work has been submitted to the IEEE for possible publication.
        Copyright may be transferred without notice, after which this
        version may no longer be accessible.}%
    }%
  \end{minipage}%
}%
\IEEEpubidadjcol

\begin{abstract}
The advancement of cooperative autonomous vehicle systems depends heavily on effective coordination between multiple agents, aiming to enhance traffic efficiency, fuel economy, and road safety. Despite these potential benefits, real-world testing of such systems remains a major challenge and is essential for validating control strategies, trajectory modeling methods, and communication robustness across diverse environments. To address this need, we introduce ConvoyNext, a scalable, modular, and extensible platform tailored for the real-world evaluation of cooperative driving behaviors.
We demonstrate the capabilities of ConvoyNext through a series of experiments involving convoys of autonomous vehicles navigating complex trajectories. These tests highlight the platform’s robustness across heterogeneous vehicle configurations and its effectiveness in assessing convoy behavior under varying communication conditions, including intentional packet loss. Our results validate ConvoyNext as a comprehensive, open-access testbed for advancing research in cooperative autonomous vehicle systems.
\end{abstract}


\section{Introduction}
\IEEEPARstart{T}{he} increasing deployment of autonomous vehicles (AVs) has paved the way for new paradigms in mobility, particularly through cooperation between multiple agents.
Cooperative driving, wherein connected vehicles share information and coordinate their actions, holds significant potential for enhancing roadway efficiency, reducing fuel consumption, and improving overall transportation safety.
Compared to isolated decision-making in single-vehicle systems, cooperative frameworks offer substantial advantages in managing dynamic traffic scenarios, especially when deployed in convoying or platooning formations.

However, developing and validating these cooperative systems in the real world remains a formidable challenge. Algorithms for coordination, trajectory planning, control, and communication must be rigorously tested under diverse and unpredictable conditions. Existing simulation environments, while useful for prototyping, often fail to capture the complexity and variability of physical deployment. Furthermore, many current testbeds lack scalability, flexibility, or support for heterogeneous vehicle types and modular control architectures.

To bridge this gap, we present ConvoyNext, a scalable and extensible testbed platform specifically designed for the real-world evaluation of cooperative autonomous vehicle systems. ConvoyNext builds upon key concepts such as modularity, controller integration, and robust communication, enabling researchers to rapidly prototype, deploy, and analyze multi-vehicle driving strategies in physical environments.

The main contributions of this work are as follows:
\begin{compactitem}
\item We introduce ConvoyNext, an open-access and scalable testbed platform for real-world cooperative driving research~\cite{Mygithub}.

\item We implement and evaluate modular trajectory tracking controllers integrated with communication-aware platooning schemes.

\item We conduct extensive experiments across multiple vehicle configurations and complex curved paths, including scenarios with controlled communication degradation.

\end{compactitem}

By offering a flexible and realistic testing environment, ConvoyNext provides a critical step toward the deployment and validation of safe and reliable cooperative autonomous vehicle systems.

\section{Related Work}
\label{sec:related}

Cooperative‐driving research spans four evaluation styles—pure \emph{simulation}, \emph{software‑in‑the‑loop} (SIL), \emph{hardware‑in‑the‑loop} (HIL), and full \emph{real‑world} testing—each providing a different fidelity/cost trade‑off.  Table~\ref{tab:related_work} summarises representative platforms in these categories and highlights the key capabilities that ConvoyNext contributes but existing testbeds largely lack.

\subsection{Simulation and SIL Testbeds}
Early‑stage Cooperative Driving Automation (CDA) algorithms are commonly developed in simulation or SIL environments.  
OpenCDA~\cite{OpenCDA} and its ROS extension~\cite{OpenCDA-ROS} couple CARLA and SUMO, supplying modular perception, planning, and merge pipelines that can be executed entirely in SIL.  
SimuSIL~\cite{kulathunga2024simusil} augments Veins with Webots and OMNeT++ to capture packet‑loss effects and eco‑driving metrics, while  
TEPLITS~\cite{TEPLITS} combines ROS, CarMaker, and a GNSS simulator for cooperative localisation studies.  
Although invaluable for prototyping, these tools remain simulation‑centric and cannot provide hardware‑validated, multi‑vehicle field results.

\subsection{HIL Platforms}
To increase realism without full road testing, HIL frameworks integrate real controllers or sensors with virtual environments.  
MicroIV~\cite{MicroIV} demonstrates low‑cost cooperative manoeuvres on miniature vehicles, and  
EI‑Drive~\cite{EI-Drive} evaluates cooperative perception with realistic communication latency models.  
These platforms, however, are limited in scale or vehicle diversity and often test only simple longitudinal controllers.

\subsection{Real‑World Testbeds}
Full‑scale deployments are rarer because of cost and safety constraints.  
FHWA’s CARMA programme validates CACC and merge logic on commercial AVs, while the vision‑tracked platform by Sheng \emph{et al.}~\cite{sheng2006cooperative} explores distributed motion planning.  
Such systems are powerful but proprietary or inflexible, making iterative academic experimentation difficult.

\subsection{Position of ConvoyNext}
ConvoyNext bridges the gap between SIL convenience and real‑world fidelity.  
It offers an \emph{open‑access}, \emph{modular} architecture that supports heterogeneous vehicles, geometric and velocity controllers, and communication‑aware coordination on both straight and curved tracks.  
By emulating packet loss and enabling rapid controller swap‑outs, ConvoyNext delivers a uniquely scalable testbed for reproducible CDA studies—capabilities that are absent or only partially met in prior work (see Table~\ref{tab:related_work}).

\begin{table}[t]
\caption{Comparison of Representative CDA Testbeds}
\label{tab:related_work}

{\centering
\resizebox{\linewidth}{!}{%
\begin{tabular}{lcccccc}
\toprule
\textbf{Platform} & \textbf{Type} & \textbf{Open} & \textbf{ROS} &
\textbf{Real Veh.} & \textbf{Comm.\ Model} & \textbf{Mod.\ Ctrl.} \\ \midrule
OpenCDA             & SIL     & \cmark & \cmark & \xmark     & \cmark & \cmark \\
OpenCDA‑ROS         & SIL     & \cmark & \cmark & Limited    & \cmark & \cmark \\
SimuSIL             & SIL     & \cmark & \cmark & \xmark     & \cmark & \xmark \\
TEPLITS             & SIL/HIL & \xmark & \cmark & \xmark     & \cmark & \xmark \\
MicroIV             & HIL     & \cmark & \xmark & Miniature  & \cmark & \xmark \\
EI‑Drive            & HIL     & \xmark & \xmark & \cmark     & \cmark & \xmark \\
CARMA               & Real    & \xmark & \xmark & Full‑scale & \cmark & Limited \\
Sheng \emph{et al.} & Real    & \xmark & \xmark & \cmark     & \cmark & \xmark \\ \midrule
\textbf{ConvoyNext} & Real    & \cmark & \cmark & \cmark     & \cmark & \cmark \\ \bottomrule
\end{tabular}}%
\par}   

\scriptsize\raggedright
Comm.\ Model = Explicit communication‑channel modelling;\,
Mod.\ Ctrl. = Modular controller integration.
\end{table}

\section{Architecture of ConvoyNext}

\begin{figure*}[t]
\centerline{\includegraphics[width=1.0\linewidth]{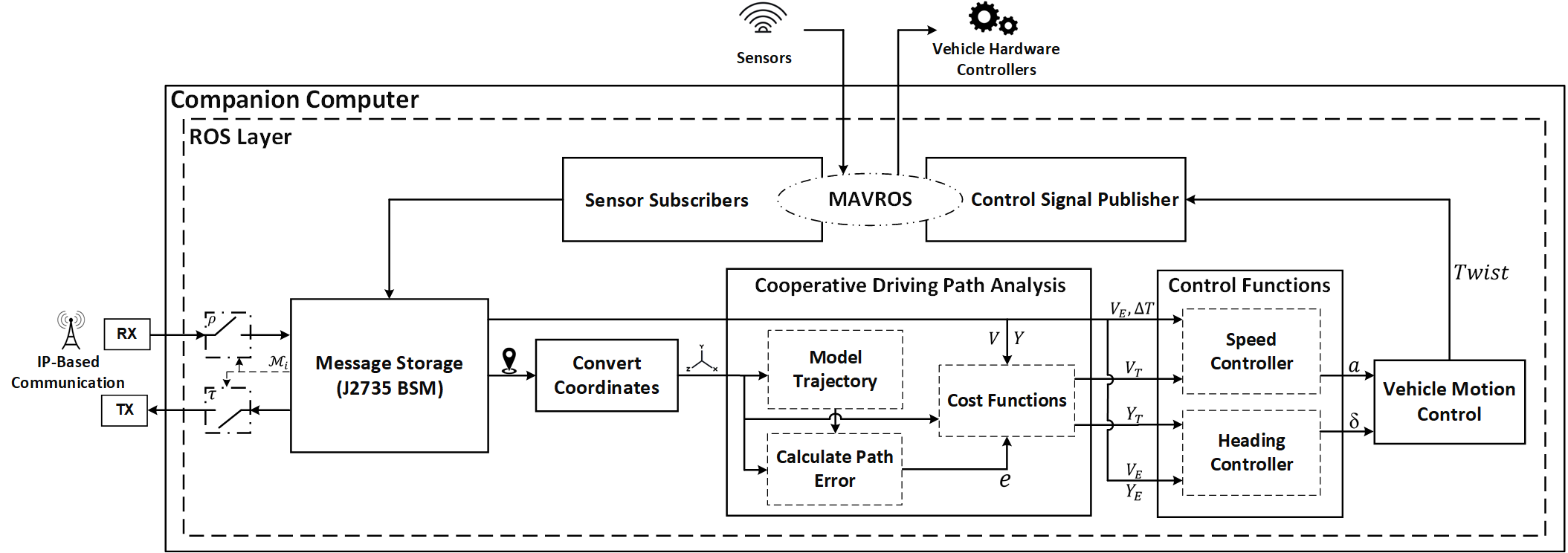}}
\caption{Block diagram of ConvoyNext Platform}
\label{block_diagram}
\end{figure*}

ConvoyNext is designed as a modular real-time architecture that enables coordinated autonomous vehicle behavior using a layered control and communication framework. As illustrated in Fig.~\ref{block_diagram}, the architecture integrates sensor data, inter-vehicle communication, control functions, and motion planning within a Robot Operating System (ROS)-based environment running on a companion computer. The system is capable of receiving Basic Safety Messages (BSMs) over IP-based communication, analyzing cooperative paths, and publishing actuation commands to the vehicle hardware via MAVROS. This structure supports a variety of controllers, communication models, and trajectory tracking methods.

\subsection{Communication Layer}

The communication layer in ConvoyNext manages the exchange and processing of inter-vehicle state information to support cooperative control. Communication occurs over standard IP-based protocols using UDP sockets, where vehicles periodically broadcast and receive state updates as Basic Safety Messages (BSMs). These messages are parsed, filtered, and stored in a structured form to support downstream decision-making.

Let $V = \{0, 1, ..., n\}$ denote the set of vehicles in the convoy, where vehicle $0$ is the designated leader. Each vehicle $i \in V$ maintains a data buffer $\mathcal{M}^i$ that stores the recent motion history of vehicles it receives messages from, including its own state. Formally, for each vehicle $j$, this buffer is expressed as:
\begin{equation}
    \mathcal{M}_j^i = \{s_j^{t-k}, s_j^{t-k+1}, ..., s_j^t\}
\end{equation}
where $s_j^t$ represents the state of vehicle $j$ at time $t$, and $k$ defines the message history window.

The system employs a reception policy function $\rho(i, j)$ that determines whether vehicle $i$ should accept and store messages from vehicle $j$:
\begin{equation}
    \rho : (i, j) \rightarrow \{0, 1\}
\end{equation}
This allows for flexible filtering topologies, including leader-only, neighbor-based, or full-broadcast modes.

On the transmission side, each vehicle evaluates whether to broadcast its current state at intervals of $\Delta t = \mathcal{B}$ seconds, using a broadcast gate defined as:
\begin{equation}
    \tau : \mathcal{M}^i \rightarrow \{0, 1\}
\end{equation}
This function can encode criteria such as time-based schedules, distance thresholds, or relevance to nearby vehicles. Based on the maintained buffer $\mathcal{M}^{i}$, each vehicle computes its cooperative motion targets.

This abstraction layer decouples the communication protocol from the control policy, enabling ConvoyNext to support a wide range of cooperative strategies, including those with intermittent communication, dynamic topologies, or priority-based message handling.

\subsection{Sensor–Command Interface in ConvoyNext}
\label{subsec:sensors_pub}

\begin{figure}[!t]
\centerline{\includegraphics[width=1.0\linewidth]{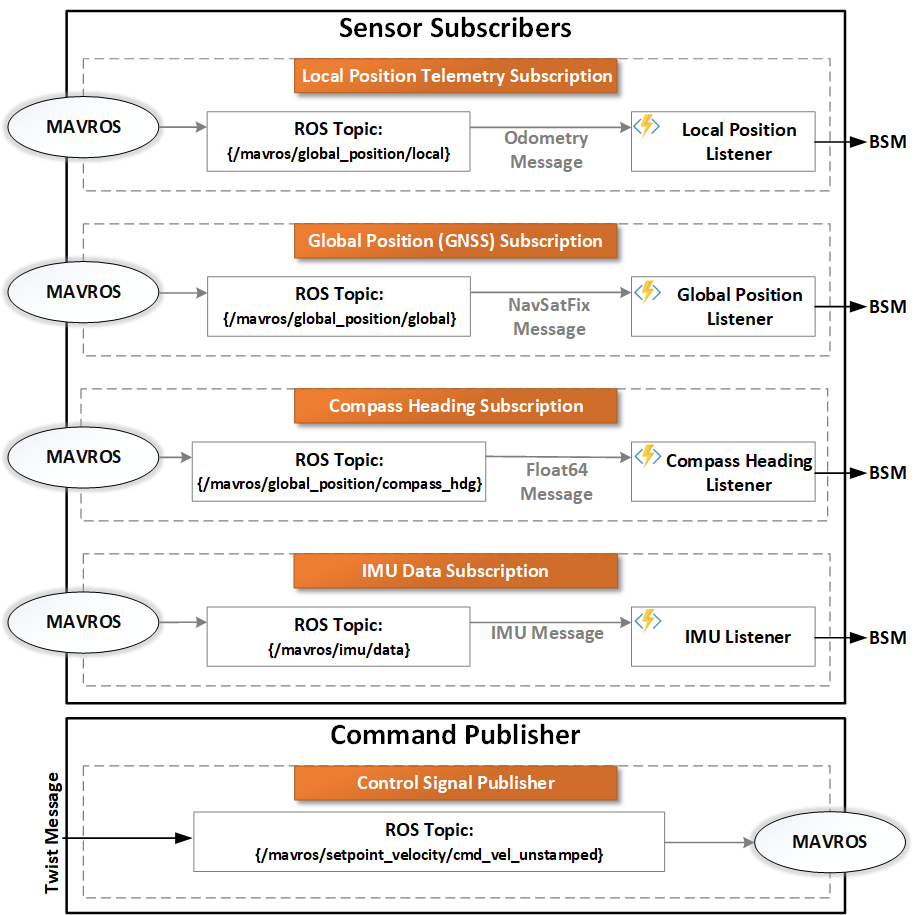}}
\caption{Subscribers and Publisher in ConvoyNext}
\label{fig:sub_pub_arch}
\end{figure}

Figure~\ref{fig:sub_pub_arch} shows the ROS–MAVROS bridge that links
(i) a configurable bank of \emph{sensor subscribers} to on‑board
sensors, and
(ii) a \emph{command publisher} that forwards high‑level control
decisions to the flight (or drive‑by‑wire) controller.
Telemetry streams—such as odometry, GNSS fixes, compass headings, and
IMU data—arrive on dedicated ROS topics and are converted into a
normalised state vector for cooperative decision making.
Conversely, a single MAVROS topic carries a \texttt{Twist} message
containing the desired longitudinal velocity and steering set‑point,
ensuring that planning outputs are executed in real time.
This bidirectional interface is fully extensible: additional sensors
or alternative actuation channels can be integrated simply by adding
new ROS topics or remapping the published command.

\subsubsection{Sensor Subscribers}
Typical telemetry sources include:
\begin{compactitem}
  \item \textbf{Local Odometry} – six‑degree‑of‑freedom pose used for
        path‑error computation.
  \item \textbf{Global GNSS Fix} – absolute position needed for
        convoy‑level spatial alignment and BSM generation.
  \item \textbf{Compass Heading} – orientation check to maintain
        trajectory alignment.
  \item \textbf{IMU} – linear acceleration and angular‑rate data for
        high‑bandwidth stabilisation.
\end{compactitem}
The subscriber list is not fixed; users may add LiDAR, camera, or
dedicated V2X links by introducing new topics without changing the
core pipeline.

\subsubsection{Command Publisher}
After cooperative planners determine the target speed
$\,v^\ast\,$ and steering angle $\,\delta^\ast$, the values are
wrapped into a \texttt{Twist} message and published via MAVROS.
The onboard controller then converts these set‑points into actuator
signals, closing the perception–planning–actuation loop with minimal
latency.

\subsubsection{Design Considerations}
The ROS–MAVROS bridge emphasises modularity and scalability.
Sensor additions or controller swaps require only topic remapping,
making ConvoyNext a versatile platform for diverse cooperative‑driving
experiments.

\subsection{Coordinate Transformation}
\label{subsec:coord_tf}

\begin{figure}[t]
    \centering
    \includegraphics[width=\linewidth]{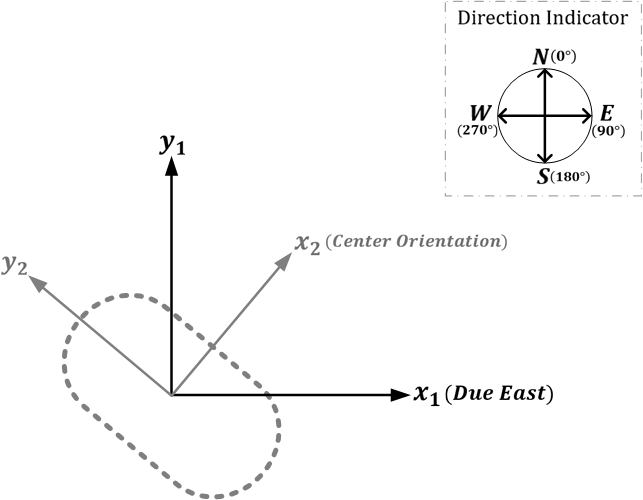}
    \caption{Local Cartesian frame definition.  
    Black axes $(x_{1},y_{1})$ are aligned with due east and north;  
    grey axes $(x_{2},y_{2})$ are rotated by the track orientation $\theta$.}
    \label{fig:coordstolocal}
\end{figure}

Position data in Basic Safety Messages (BSMs) arrive as latitude–longitude pairs.  
For convoy–level planning these global coordinates must be mapped into a common local Cartesian frame.  
Figure~\ref{fig:coordstolocal} illustrates the adopted frame: $x_{1}$ is aligned with geographic east (\ang{90}), while the grey axes $x_{2}$–$y_{2}$ represent the same basis after a user‑defined rotation~$\theta$ that matches the track’s principal orientation.  
The origin is an arbitrary reference point—e.g., the geometric centre of the test track, a roadside unit, or any surveyed marker—so the framework can be deployed at different sites without re‑calibration.

\vspace{2pt}
\noindent\textbf{Projection.}
Let $(\phi,\lambda)$ denote latitude and longitude in radians, and let $(\phi_{c},\lambda_{c})$ denote the reference point.  
Using the equirectangular approximation, global coordinates are converted to metre‑scaled east–north offsets:
\[
  x = R_{\oplus}\cos\!\bigl(\tfrac{\phi+\phi_{c}}{2}\bigr)\,(\lambda-\lambda_{c}),
  \qquad
  y = R_{\oplus}\,(\phi-\phi_{c}),
\]
where $R_{\oplus}$ is Earth’s mean radius.

\vspace{2pt}
\noindent\textbf{Frame rotation.}
To align the local $y$‑axis with the track heading, we rotate $(x,y)$ by the angle $\theta$ (measured clockwise from due east):
\[
  \begin{bmatrix} q_{x}\\ q_{y} \end{bmatrix}
    =
  \begin{bmatrix}
    \cos\theta & -\sin\theta\\
    \sin\theta &  \cos\theta
  \end{bmatrix}
  \begin{bmatrix} x\\ y \end{bmatrix}.
\]
The resulting $(q_{x},q_{y})$ coordinates are used throughout ConvoyNext’s planning and control layers, greatly simplifying trajectory‑error computation and controller tuning.

\subsection{Co‑operative Driving Path Analysis}
\label{subsec:path_analysis}

This block combines the most recent Basic Safety Messages
$\mathcal{M}_{i}$ with the local position stream to generate concrete motion targets for the ego vehicle.  Although the reference implementation
realises a platooning policy, each grey dashed sub‑module can be
replaced to support other co‑operative behaviours such as CACC,
merge sequencing, or intersection negotiation.

\paragraph*{1) Trajectory modelling}
\textit{Model Trajectory} constructs a short‑horizon prediction of
the immediately preceding vehicle.  A straight‑line fit suffices
for highway experiments, whereas a quadratic fit is used on
closed tracks with notable curvature.  Any model that outputs a
parametric path $\gamma(s)$ and its tangent heading can be
plugged‑in.  Here
$\gamma(s)\!:\![0,S]\!\rightarrow\!\mathbb{R}^{2}$ denotes the predicted
centre‑line of the predecessor, and
$\theta(s)=\tan^{-1}\!\bigl(\dot{y}(s),\,\dot{x}(s)\bigr)$ is its
instantaneous heading.  Thus straight lines, splines, clothoids,
or learned predictors can all be substituted without modifying
later stages of the pipeline.

\paragraph*{2) Path‑error computation}
Given the predicted trajectory and the ego vehicle’s current
Cartesian pose $(q_x,q_y)$, the
\textit{Calculate‑Path‑Error} block returns  
(i) the signed cross‑track error $e$ and  
(ii) a preview point $\bigl(x_{T},y_{T}\bigr)$ chosen either

\begin{itemize}
  \item at a fixed geometric separation $d$ (distance‑gap policy), or
  \item at a bumper‑to‑bumper spacing $d = v_{T}\,\Delta T$  
        (time‑gap policy with headway $\Delta T$),
\end{itemize}

where $v_{T}$ is the predicted speed of the predecessor.  
Both policies are supported; the choice is made by the
controller configuration.

\paragraph*{3) Cost synthesis}
The default objective implemented in Algorithm 1 is the
sum of Euclidean position errors between the ego vehicle’s predicted
future pose \((x_{e}^{f},y_{e}^{f})\) and the \emph{goal} pose
\((x_{e}^{g},y_{e}^{g})\) that realises the desired following distance
to each predecessor:
\[
  J \;=\;
  \sum_{j\in\mathcal{P}}
  \sqrt{(x_{e}^{g,j}-x_{e}^{f})^{2}+(y_{e}^{g,j}-y_{e}^{f})^{2}},
\]
where \(\mathcal{P}\) is the set of vehicles the ego must track.
Researchers may override this function to incorporate speed,
heading, fuel‑use, or any other penalty term.

\paragraph*{4) Target generation}
A constrained optimiser searches the
space of admissible longitudinal velocities
$v_{E}\!\in[v_{\min},v_{\max}]$ and heading offsets
$\Delta\theta\!\in[-\theta_{\max},\theta_{\max}]$ to minimise~$J$.
The resulting pair \((v_{T},\theta_{T})\) is placed on the shared bus
\(V_{T},Y_{T}\) in Fig.~\ref{block_diagram} and consumed by the
\textit{Control Functions} block.  Because both the objective and
the solver are user‑selectable, users can implement
gradient‑free heuristics, MPC‑style quadratic programmes, or
learning‑based policies simply by subclassing the supplied
\texttt{Control} class.

\paragraph*{5) Extensibility}
The module cleanly separates \emph{what} co‑operative objective
is pursued from \emph{how} low‑level actuation is executed,
making the platform suitable for comparative studies across a
wide range of multi‑vehicle policies.

\begin{algorithm}[t]
\caption{Default trajectory--cost evaluation (platooning)}
\begin{algorithmic}[1]
\State \textbf{Input:} $v_{\text{ego,input}},\, \text{yaw}_{\text{ego,input}}$
\State $(x_{\text{ego}}, y_{\text{ego}}) \gets \text{ConvertToLocal}(\text{state.latitude},\, \text{state.longitude})$
\State $total\_cost \gets 0$
\For{each target in \texttt{car\_positions}}
    \State $T \gets \text{LatestRecord}(target)$
    \State $(x_{\text{target}}, y_{\text{target}}) \gets \text{ConvertToLocal}(T.\text{latitude},\, T.\text{longitude})$
    \State $\text{yaw}_{\text{target}} \gets \text{DegreesToRadians}(T.\text{heading})$
    \State $v_{\text{target}} \gets T.\text{speed}$
    \State $position \gets car\_number - T.\text{event\_flags}[\texttt{'car'}]$
    \State $GFD \gets follow\_distance \times position$
    \State $x_{\text{target,future}} \gets x_{\text{target}} + (v_{\text{target}} \cdot \sin(\text{yaw}_{\text{target}}) \cdot BcastInt)$
    \State $y_{\text{target,future}} \gets y_{\text{target}} + (v_{\text{target}} \cdot \cos(\text{yaw}_{\text{target}}) \cdot BcastInt)$
    \State $x_{\text{ego,future}} \gets x_{\text{ego}} + (v_{\text{ego}} \cdot \sin(\text{yaw}_{\text{ego}}) \cdot BcastInt)$
    \State $y_{\text{ego,future}} \gets y_{\text{ego}} + (v_{\text{ego}} \cdot \cos(\text{yaw}_{\text{ego}}) \cdot BcastInt)$
    \State $x_{\text{ego,goal}} \gets x_{\text{target,future}} - (GFD \cdot \sin(\text{yaw}_{\text{target}}))$
    \State $y_{\text{ego,goal}} \gets y_{\text{target,future}} - (GFD \cdot \cos(\text{yaw}_{\text{target}}))$
    \State $cost \gets \sqrt{(x_{\text{ego,goal}} - x_{\text{ego,future}})^2 + (y_{\text{ego,goal}} - y_{\text{ego,future}})^2}$
    \State $total\_cost \gets total\_cost + cost$
\EndFor
\State \textbf{Output:} $total\_cost$
\end{algorithmic}
\end{algorithm}

\subsection{Low‑Level Control}
\label{subsec:control}

The Control Functions block converts the
high‑level set‑points \((v^{\ast},\theta^{\ast})\) into throttle/steer
commands understood by the vehicle interface.  Any
speed‑holding and heading‑holding laws—PID, MPC, LQR or
data‑driven—can be dropped in; the framework simply expects a
routine that maps
\(\bigl(v^{\ast},\theta^{\ast},\text{state}\bigr)\rightarrow
(\text{longitudinal\;accel},\text{steer\;angle})\).
Because sensing, network I/O and logging are handled
elsewhere, new controllers can be evaluated by overriding a
single callback without touching the rest of the pipeline.

\begin{figure}[t]
    \centering
    \includegraphics[width=\linewidth]{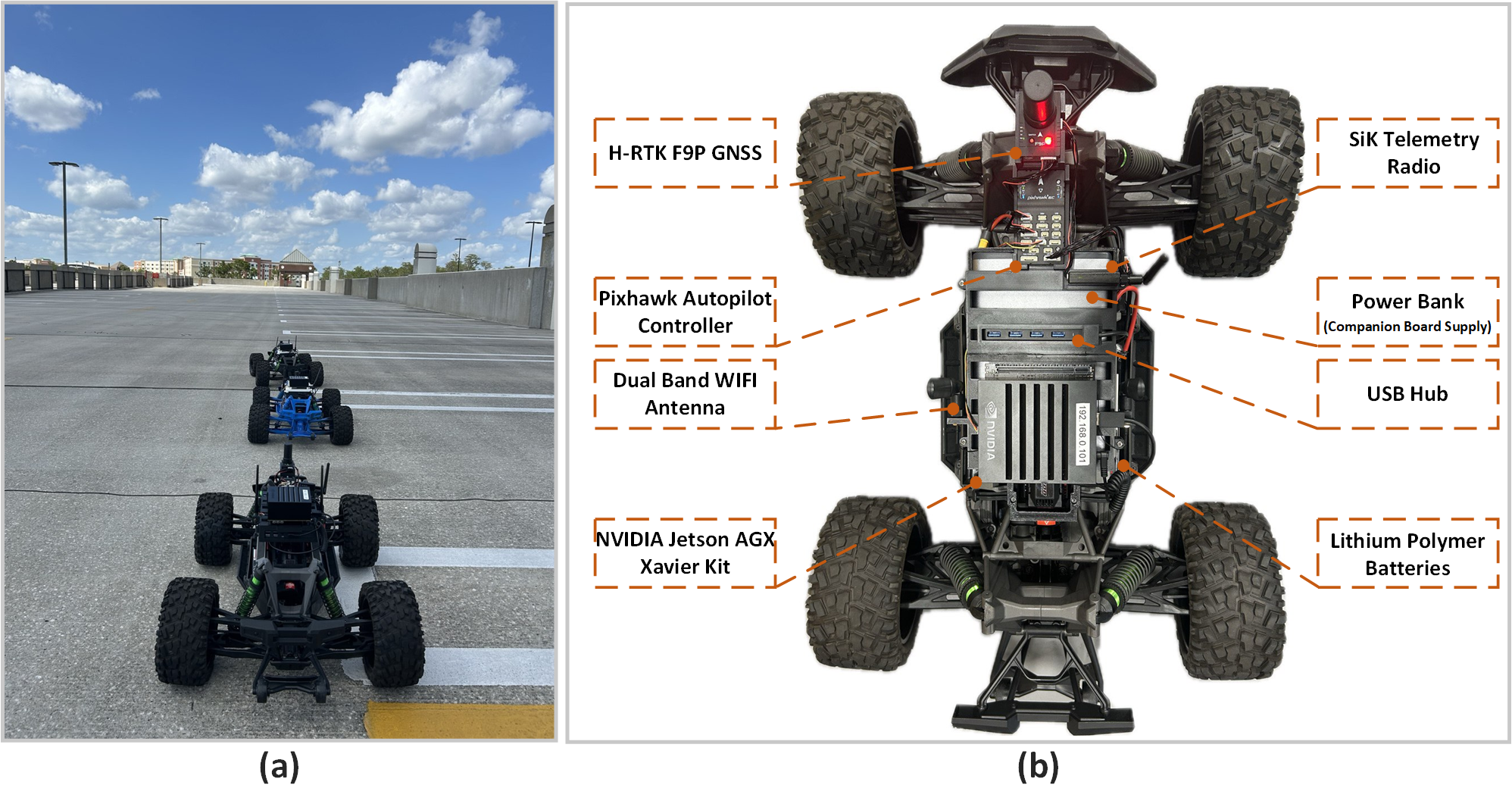}
    \caption{%
  (a) Three‑vehicle convoy on the test site;
  (b) hardware layout of the autonomous buggy.}
    \label{fig:testbed}
\end{figure}

\section{Experiments}

All trials were performed with a convoy of three \(1{:}6\)-scale electric vehicles equipped with GNSS, Wi‑Fi V2V radios, and a Pixhawk‑based drive‑by‑wire stack (Fig.~\ref{fig:testbed}). The ConvoyNext software ran on each vehicle and was configured for

\begin{itemize}
\item \emph{all‑predecessor following}: the information‑flow
      topology returns \(\tau=1\) for every broadcast, and
      \(\rho=1\) only if the sender is an upstream neighbour;
\item time‑triggered broadcasting at
      \(f_{\text{bcast}}=\SI{10}{\hertz}\) with packet‑drop probabilities
      \(\{0,\,0.1,\,0.4,\,0.5\}\) to emulate congested channels;
\item a non‑model‑based planner followed by a Stanley
      lateral controller and a PID speed controller.
\end{itemize}

\begin{figure}[t]
  \centering
  \includegraphics[width=0.85\linewidth]{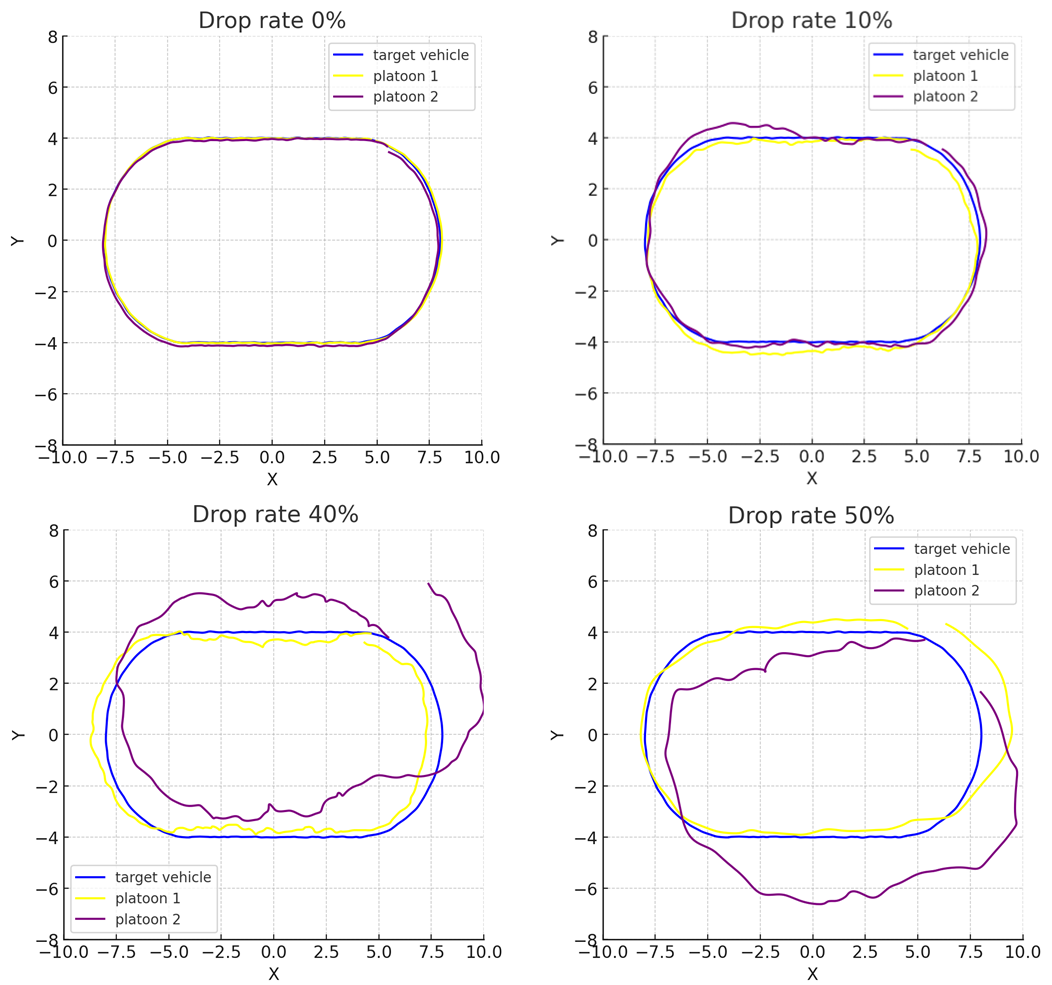}
  \caption{Plan-View Trajectories Under V2V Packet Drop Rates.}
  \label{fig:result}
\end{figure}

The leader traversed an \(\SI{8}{\metre}\times\SI{4}{\metre}\) oval.
For the first half of every lap its speed was
\SI{1}{\metre\per\second}; during the second half it was increased
to \SI{2}{\metre\per\second}.  The followers attempted to match
this time‑varying velocity while maintaining the required
inter‑vehicle gaps, so any packet loss immediately manifested
as a lateral or longitudinal tracking error (Fig.~\ref{fig:result}).

Fig. \ref{fig:result} and Table \ref{tab:metrics} show clear degradation with increased packet loss. At 0\% drop, vehicles maintain near-identical paths and spacing errors below one vehicle length (\SI{8}{\centi\metre}). A mild loss (10\%) slightly widens the paths and doubles the gap error, yet control remains robust. Above 40\% loss, oscillations occur, velocity synchronization worsens, and errors grow significantly. At 50\%, instability emerges—followers overshoot corners and velocity differences peak at \SI{0.4}{\metre\per\second}, underscoring the need for reliable V2V communication.

\section{Conclusion}
ConvoyNext offers a reliable, scalable solution for real-world testing of cooperative autonomous vehicle systems. Its successful deployment across varied scenarios and communication conditions confirms its value as a comprehensive platform for advancing research and development in this field.

\begin{table}[ht]
  \centering
  
  \caption{Effect of V2V packet loss on closed‑loop platooning
           (95\textsuperscript{th}\,percentile over five laps).}
  \label{tab:metrics}
  \begin{tabular}{@{}lcc@{}}
    \toprule
    \textbf{Packet‑drop rate} &
    \textbf{Platooning error}$^{\dagger}$  &
    \textbf{Speed difference}$^{\ddagger}$ \\[-0.2em]
    & (cm) & (m\,s$^{-1}$) \\
    \midrule
     0\,\%  &  8  & 0.1   \\
    10\,\%  & 22  & 0.15   \\
    40\,\%  &  146  & 0.34   \\
    50\,\%  & 188  & 0.4 \\
    \bottomrule
  \end{tabular}
  \vspace{0.3em}
  \begin{minipage}{0.85\linewidth}
  \tiny
  $^{\dagger}$ Absolute gap error w.r.t.\ the \SI{20}{cm} set‑point. \\
  $^{\ddagger}$ Instantaneous \(\max(v) - \min(v)\) across the three vehicles.
  \end{minipage}
\end{table}

\pagebreak
\newpage

\bibliographystyle{IEEEbib}

\end{document}